\documentclass[cameraready]{Interspeech}

\title{Content is What Remains: Invariant Speech Tokenization from Parallel Utterances}

\author[affiliation={1}, correspondingauthor]{Laurin}{Wagner}
\author[affiliation={1}]{Bernhard}{Thallinger}
\author[affiliation={1}]{Miroslav}{Stankovic}
\author[affiliation={1}]{Mario}{Zusag}

\address{
    $^1$ nyra labs, Austria
}

\email{lwagner@nyra-labs.com, bthallinger@nyra-labs.com, mzusag@nyra-labs.com}


\usepackage{comment}
\usepackage{tabularx}
\usepackage{booktabs}
\usepackage{makecell}
\usepackage{fontawesome5}
\usepackage{amsmath}

\begin{document}

\maketitle

\begin{abstract}
Discrete speech tokenizers aim to disentangle semantic from acoustic information, yet targets from self-supervised learning (SSL) models like HuBERT retain non-linguistic variation: speaker identity, prosody, and channel conditions leak into tokens, inflating entropy. Our insight: when enough speakers utter the same words under varying conditions, linguistic content is the only shared factor. We propose \textsc{PINT} (Parallel INvariant Tokenization), fine-tuning an SSL encoder with alignment losses across parallel utterances and augmentations to distill this shared residual. \textsc{PINT} collapses identical words onto consistent token sequences, drastically reducing conditional entropy. Unlike ASR text, \textsc{PINT} tokens preserve frame-level temporal grounding and serve as drop-in semantic targets for audio codecs. Experiments show a $98.7\%$ relative reduction in speaker probe accuracy ($93.1\%\!\to\!1.2\%$), $42\%$ lower ABX error rate, and $27$--$30\%$ lower LM perplexity versus baselines, confirming that the right invariance is key to efficient learning.
\footnote{Models: \url{https://github.com/nyrahealth/PINT}}
\end{abstract}

\section{Introduction}
\label{sec:intro}

Subword tokenization in NLP succeeds partly because discretization that respects a signal’s latent structure yields sequences that are highly compressible and predictable for autoregressive models. Discrete speech tokenization extends this premise to audio, supporting LM-based speech generation, dialogue, and cross-modal reasoning~\cite{borsos2023audiolm,zhang2023speechgpt,defossez2024moshi,wang2023viola}. In contemporary speech-codecs and speech generation systems, tokens often appear in a factorized representation (per timestep/frame): \emph{semantic} tokens should capture linguistic content, while \emph{acoustic} residual layers model speaker, prosody, and other signal details~\cite{zhang2024speechtokenizer,wu2023audiodec, defossez2024moshi}.

\noindent\textbf{The nuisance-leakage problem.}
Semantic tokens in tokenizers commonly discretize intermediate SSL features, typically HuBERT~\cite{hsu2021hubert} or WavLM~\cite{chen2022wavlm}, via distillation~\cite{zhang2024speechtokenizer,defossez2024moshi} into codebooks or dual encoders where one encodes semantics and the other fills in the gaps ~\cite{xy-tokenizer}. However, these encoders leak non-linguistic variation: speaker identity remains recoverable~\cite{qian2022contentvec,yeh2024completeness}, and token sequences can change by $>{}40\%$ under acoustic perturbations alone~\cite{gat2023augmentation}, a brittleness that has been found to correlate with downstream degradation~\cite{vashishth2024stab,chang2025dcspin,zhong2025accent}.\newline
Formally, let $c$ denote linguistic content and $\mathbf{z}$ the token sequence. When different realizations of the same $c$ yield different $\mathbf{z}$, the conditional entropy $H(\mathbf{z}\mid c)$ is bounded away from zero by non-linguistic variation. This leakage hinders both sequence compressibility and applications that benefit from disentangled content and style representations such as accent conversion~\cite{zhong2025accent}, voice conversion~\cite{qian2022contentvec}, and expressive synthesis~\cite{zhang2025vevo}, where downstream models are forced to undo entanglement that a cleaner representation would have prevented. A cleanly disentangled representation would instead allow models to manipulate style independently of content.

\noindent\textbf{Related work.}
ContentVec~\cite{qian2022contentvec} and DC-Spin~\cite{chang2025dcspin} pursue speaker invariance via contrastive or clustering objectives; Gat et al.~\cite{gat2023augmentation} and NAST~\cite{messica2024nast} target augmentation robustness; StableToken~\cite{song2025stabletoken} achieves noise invariance through multi-branch voting but does not address speaker or prosodic variation.
On the codec side, SpeechTokenizer~\cite{zhang2024speechtokenizer}, Moshi/Mimi~\cite{defossez2024moshi}, DualCodec~\cite{dualcodec}, XY-Tokenizer~\cite{xy}, and SAC~\cite{sac} each improve semantic–acoustic disentanglement through distillation, dual encoders, or split-stream quantization. Yet all inherit features (HuBERT, WavLM, Whisper \cite{whisper} or w2v-BERT \cite{w2vBert}) as semantic targets without addressing the nuisance leakage already present in those representations. PINT is orthogonal: rather than redesigning codec architectures, it cleans the encoder upstream, producing invariant representations that serve as drop-in targets for any of these frameworks.

\noindent\textbf{PINT: formulation and approach.}
We argue that a token is only truly semantic if it satisfies a dual criterion: \emph{content capture}: it carries sufficient phonetic information and \emph{nuisance invariance:} it discards everything else. nuisance invariance drives $H(\mathbf{z}\mid c)\!\to\!0$, ensuring identical content maps to identical token sequences. This allows run-length encoding (RLE) and byte pair encoding (BPE) to collapse redundant frames, drastically improving bit rate efficiency and predictability for autoregressive modeling, since the remaining entropy reflects genuine linguistic structure discarding other factors.

We introduce PINT to operationalize this through parallel data: when multiple speakers utter the same sentence under different conditions, linguistic content is the only shared factor. By fine-tuning a HuBERT encoder with parallel-data alignment losses and aggressive augmentations, PINT distills this shared residual, leaving content as the only consistent signal. Unlike ASR-derived text tokens, PINT preserves frame-level temporal grounding facilitating integration into modern audio codec architectures.

The payoff is measurable and model-agnostic: an identical 85\,M-parameter decoder-only transformer achieves \textbf{27--30\,\% lower perplexity} on PINT tokens than on HuBERT or WavLM tokens, reaching WavLM's final perplexity in 23$\times$ fewer steps - evidence that invariance, not architecture, is the binding constraint on token quality.

\noindent\textbf{Contributions.}

(1)~We formalize semantic speech tokenization through a dual criterion of content capture and nuisance invariance, showing that nuisance leakage directly inflates conditional entropy and hurts compressibility.
(2)~We propose PINT, a training framework leveraging parallel data and augmentation to jointly optimize continuous and discrete representations, satisfying both criteria simultaneously.
(3)~We show that PINT yields highly compressible content tokens, reducing speaker probe accuracy by 98.7\,\% (93.1\,\%\,$\to$\,1.2\,\%), improving ABX discriminability by 42\,\%, and lowering LM perplexity by 27--30\,\% vs.\ HuBERT/WavLM (test perplexity 1.95 vs.\ 2.78/2.67).

\section{Method}
\label{sec:method}

\subsection{Data and Parallel Supervision}
\label{ssec:data}

Training and evaluation draw on four categories of English speech, summarized in Table~\ref{tab:datasets}.
\textbf{True-parallel corpora} (ARCTIC, CHAINS, CSTR-VCTK, EnDialects, ESD, TIMIT) provide multiple recordings of the same transcripts across speakers, accents, speaking styles, and emotional conditions, yielding naturally aligned utterance groups that span a wide range of variability.
\textbf{Non-parallel corpora} (LibriSpeech 960\,h, Tedlium-3 374\,h) extend linguistic and acoustic coverage; since they lack shared transcripts, pseudo-parallel pairs are formed on-the-fly via stochastic augmentations like additive noise (white, brown, pink), reverberation, speed/pitch perturbation, and channel distortion applied to the same utterance within a batch. Further we synthesize multiple parallel samples for these datasets using Kokoro~\cite{kokoro}
\textbf{Evaluation-only}: RAVDESS~\cite{ravdess} (emotion probes) and 6k hours clean subset of Librilight for LM training. 
\textbf{Noise}: We treat noise~\cite{noise} as parallel data with empty transcripts during training.
All corpora are force-aligned with the Montreal Forced Aligner (MFA)~\cite{mfa} for word-level timestamps. Evaluations use data strictly excluded from training: CSTR-VCTK holds out 10\,\% of speakers; TIMIT and LibriSpeech use standard non-overlapping partitions; for corpora without official splits, all utterances sharing a normalized transcript are assigned to the same split to prevent transcript leakage.\footnote{The Use column in Table~\ref{tab:datasets} codes each corpus's role: T=Stage~A training, KM=$k$-means fitting, I=invariance evaluation, LM=language model.}

\begin{table}[t]
\centering
\footnotesize
\caption{Dataset statistics and parallel properties. Hours: total audio length used; \#S: parallel samples; \#G: parallel groups. Use codes: T=Stage~A; KM=$k$-means; I=invariance evaluation; LM=language model.}
\label{tab:datasets}
\begin{tabular}{lrrrp{1.2cm}l}
\toprule
Dataset & Hours & \#S  & \#G  & Parallel properties  & Use \\
\midrule
ARCTIC~\cite{arctic}            & 13.7  & $15$k   & 1134  & accents/ dialects  & T  \\
CHAINS~\cite{chains}            & 9.6   & $6$k    & 56    & accents, speed, style  & T \\
CSTR-VCTK~\cite{cstr}          & 82.5  & $88$k   & 13615 & accents/ dialects  & T  I \\
EnDialects~\cite{english-dialects}  & 30.4  & $17$k   & 2273  & accents/ dialects  & T \\
ESD (EN)~\cite{esd}             & 13.3  & $17$k   & 383   & emotions  & T\\
RAVDESS~\cite{ravdess}          &  1.4  & 720     &   2   & emotions & I \\
SynSpeech ~\cite{synspeech}                       & 196.6 & $109$k  & 109   &      & T \\
TIMIT~\cite{timit}              & 3.7   & $4.4$k  & 452   &      & T KM \\
Noise~\cite{noise}                            & 30.0  & $10$k   &   1   &       & T I \\
\midrule
LibriSpeech~\cite{librispeech}  & 120.4 & $73$k   & ---   & ---  & T I\\
Tedlium-3~\cite{tedlium3}      & 374.8 & $244$k  & ---   & ---  & T \\
\midrule
LibriLight~\cite{librilight}    &   6000    &         & ---   & ---  & LM \\
\bottomrule
\end{tabular}
\end{table}

\subsection{Model Overview}
\label{ssec:model}
PINT starts from HuBERT-base~\cite{hsu2021hubert}. Given a waveform $\mathbf{x}$, the encoder produces continuous frame-level features that serve as input to a decoder branch predicting phoneme sequences via cross-entropy (CE). In Stage~B, an additional id-sequence head outputs logits over $K{=}200$ discrete codes plus a blank symbol for connectionist temporal classification (CTC) training~\cite{graves2006ctc}.

\subsection{Stage A: Parallel Invariance Training}
\label{ssec:stageA}

Mini-batches contain utterances indexed by $b\in\{1,\dots,B\}$, where $g_b$ denotes the transcript-group index.

\noindent\textbf{(1) Sequence-level soft dynamic time warping (DTW) loss.}
Let $\mathbf{H}_b=[\mathbf{h}_{b,1},\dots,\mathbf{h}_{b,T_b}]\in\mathbb{R}^{T_b\times D}$ denote encoder frame features for utterance $b$, where $T_b$ is the number of frames and $D$ is the feature dimension. For each positive parallel pair $(i,j)$ with $g_i{=}g_j$, let $\mathcal{P}$ be the set of all such pairs in the mini-batch. Soft-DTW~\cite{cuturi2017softdtw} computes a distance $d^{\mathrm{sdtw}}_{ij}$ between $\mathbf{H}_i$ and $\mathbf{H}_j$, normalized by the maximum sequence length:
\begin{equation}
\label{eq:sdtw}
\mathcal{L}_{\mathrm{sdtw}}=\frac{1}{|\mathcal{P}|}\sum_{(i,j)\in\mathcal{P}} \frac{d^{\mathrm{sdtw}}_{ij}}{\max(T_i,T_j)}.
\end{equation}
Since parallel pairs share only linguistic content while differing in speaker, style, recording conditions etc. minimizing this loss forces the encoder to output consistent representations for content and suppress nuisance variation.

\noindent\textbf{(2) Word-level contrastive loss.}
Using word boundaries from forced alignment, we compute word representations by averaging encoder frames over each word span:
$\mathbf{v}_{b,m} = \frac{1}{|\mathcal{T}_{b,m}|}\sum_{t \in \mathcal{T}_{b,m}} \mathbf{h}_{b,t}$.
Positive pairs $\mathcal{V}^{+}$ consist of the same word position across parallel utterances; negative pairs $\mathcal{V}^{-}$ consist of words whose phoneme sets have Jaccard similarity below a threshold~$\tau_{\mathrm{neg}}$.\footnote{Jaccard-based gating avoids contradictory gradients from phonetically similar negatives.}
The loss attracts same-word representations and repels different ones:
\begin{equation}
\label{eq:word}
\mathcal{L}_{\mathrm{word}} =
1 - \overline{\cos}(\mathcal{V}^{+})
\;+\;\lambda_{\mathrm{neg}}\,
\overline{\cos}(\mathcal{V}^{-}),
\end{equation}
where $\overline{\cos}(\cdot)$ is the duration-weighted mean cosine similarity.
This attraction-repulsion prevents the trivial collapse that the sDTW loss alone would suffer from and further encourages representations that are aligned on the word level.

\noindent\textbf{(3) Decoder CE loss.}
A two-layer Transformer decoder trained with teacher-forced CE over phoneme targets from transcript phonemization, attending to the encoder output via cross-attention ensures \emph{content capture}.\newline
The Stage~A objective combines all three losses (all $\lambda_*$ are scalar weights):
\begin{equation}
\label{eq:loss_a}
\mathcal{L}_{\mathrm{A}}=
\lambda_{\mathrm{sdtw}}\mathcal{L}_{\mathrm{sdtw}}+
\lambda_{\mathrm{word}}\mathcal{L}_{\mathrm{word}}+
\lambda_{\mathrm{ce}}\mathcal{L}_{\mathrm{ce}}.
\end{equation}

We set $\lambda_{\mathrm{sdtw}}=.5$, $\lambda_{\mathrm{word}}=2$, $\lambda_{\mathrm{neg}}=1$, and $\lambda_{\mathrm{ce}}=10$.
\subsection{Stage B: Discrete Token Training}
\label{ssec:stageB}

Stage~A captures content with invariant continuous embeddings, yet converting them to tokens via a post-hoc $k$-means codebook does not guarantee sequence-level optimality since the clustering objective is agnostic to sequential consistency across parallel utterances. Stage~B addresses this via end-to-end optimization that directly targets discrete sequence consistency while jointly fine-tuning all components, enforcing invariance in both continuous and discrete form:  Following Lee et~al.~\cite{lee2022textless}, we attach a linear layer that maps encoder frames to logits over a discrete unit vocabulary. For each transcript group, the teacher (an EMA copy of the student) processes an anchor utterance (synthetic for synth-aug), obtains token IDs by framewise argmax decoding of these logits, and then applies deduplication to form a discrete target sequence. This yields a common reference sequence per group. The student is trained with CTC to align every utterance in the group to this shared target. Because CTC permits many-to-one alignments, variation in duration and speaking rate is absorbed without penalty, pushing all speakers toward the same discrete output. Unlike Lee et~al., who freeze the encoder, we train end-to-end via a student-teacher framework~\cite{tarvainen2017meanteacher, data2vec}. Joint updates allow the semantic encoder to co-adapt with the discrete partition; (blanks are disallowed, as deduplication makes them unnecessary).
$\mathcal{L}_{\mathrm{marg}}$ promotes a uniform argmax distribution on the batch level to avoid collapse, while $\mathcal{L}_{\mathrm{orth}}$ discourages correlations among its prototype vectors to improve code separability. We downweight the discrete losses relative to the Stage~A terms ($\lambda_{\mathrm{id}}{=}0.6$, $\lambda_{\mathrm{m},\mathrm{o}}{=}0.3$) so that continuous invariance training remains the dominant learning signal.

\section{Experiments}
\label{sec:experiments}

We evaluate PINT along three axes that directly test the dual criterion: content capture~(\S\ref{ssec:content}) and invariance and robustness~(\S\ref{ssec:invariance}). Further we evaluate the downstream compression benefit unlocked by the above. Baselines are HuBERT-base layer~9 and WavLM-base layer~12 since these layers have been found to correspond most closly to phonemes  ~\cite{hsu2021hubert, chen2022wavlm}. Unless noted, discrete tokens use $k$-means ($k{=}200$) fit on TIMIT for baselines. 
PINT remains strong at far smaller vocabularies, while the baselines degrade more quickly and saturate around 200; we therefore use K=200 for all systems for direct comparability. Ablations: \emph{dec-AR} uses exclusively auto-regressive phoneme decoding as loss; \emph{dec-CTC} uses exclusively a CTC loss for training; \emph{synth-only} uses only synthetic data from Kokoro \cite{synspeech} and all losses; \emph{synth-aug} uses all losses and real data, but all parallel data for the real data is synthetically generated meaning real samples only interact with synthetic ones in DTW and Word level losses but never with other real ones; \emph{w/o noise} drops noise augmentation and noise as parallel data.

\subsection{Content Capture}
\label{ssec:content}

We measure content capture via ASR performance, phoneme discriminability (ABX), and phone-normalised mutual information (PNMI).
\textbf{Word error rate (WER) / character error rate (CER)}: following ~\cite{zhang2024speechtokenizer} we train a 2-layer bidirectional long short-term memory (BLSTM) network (hidden dimension~1024) with CTC on LibriSpeech train-clean-360, evaluated on test-clean; in continuous mode the BLSTM receives frame features directly, in discrete mode a k means model maps to ids or in PINT's case the argmax logit output of the last linear layer.
\textbf{ABX}~\cite{schatz2013abx}: minimal-pair phonetic discriminability on LibriSpeech dev-clean via the \texttt{zrc\_abx2} toolkit (20\,ms frame rate); we report within-speaker and across-speaker error (lower is better).
\textbf{PNMI} as defined in ~\cite{hsu2021hubert}.
Results are shown in Table~\ref{tab:asr}.
\begin{table}[t]
\centering
\footnotesize
\setlength{\tabcolsep}{3pt}
\caption{Content capture on LibriSpeech. Each cell shows \textit{cts / dis}: continuous vs.\ $k$-means discrete ($k{=}200$). $\downarrow$~lower is better; $\uparrow$~higher is better. \textbf{Bold}: best per sub-column.}
\label{tab:asr}
\begin{tabular}{lccccc}
\toprule
& \multicolumn{2}{c}{cts / dis} & & \multicolumn{2}{c}{ABX $\downarrow$} \\
\cmidrule(lr){2-3} \cmidrule(lr){5-6}
Model & CER & WER & PNMI$\uparrow$ & wit. & acr. \\
\midrule
\multicolumn{6}{l}{\textit{Baselines}} \\
HuBERT           & 4.33 / 7.55          & 10.99 / 21.37          & 0.79          & 0.055          & 0.066 \\
WavLM            & 4.03 / 6.40          & 11.53 / 18.42          & 0.81 & 0.047          & 0.059 \\
\midrule
PINT(ours)  & 3.84 / 4.65 & 9.79 / 12.13  & 0.78          & 0.040          & 0.042 \\
\midrule
\multicolumn{6}{l}{\textit{Ablations}} \\
dec-AR           & 3.90 / 6.00 & 10.08 / 17.93 & 0.78  & 0.076 & 0.086 \\
synth-only       & 5.87 / 7.37 & 14.95 / 18.55 & 0.75  & 0.055 & 0.062 \\
w/o noise        & 3.80 / 4.75 & 9.71 / 12.52  & 0.78  & 0.043 & 0.050 \\
dec-CTC          & 4.20 / 4.73 &  / 10.77 / 12.01 & 0.77 & 0.073 & 0.086    \\
synth-aug        & 3.79 / 4.67 & 9.91 / 12.41  & 0.78  & 0.042 & 0.044 \\
\bottomrule
\end{tabular}
\end{table}
\newline PINT outperforms both baselines on CER and WER in continuous and discrete mode while achieving substantially better ABX discriminability (0.040 vs.\ 0.042 across-speaker), demonstrating that our invariance training does not sacrifice content. Among the ablations, \emph{synth-only} degrades both ASR and ABX, showing that real data is essential for real world performance; \emph{dec-CTC} and \emph{dec-AR} underperform on ABX, confirming that the full multi-loss training is beneficial for phoneme discriminability. Most importantly \emph{synth-aug} performs very well. Since the required parallelism can be generated synthetically, extending PINT to languages without parallel human corpora appears feasible.


\subsection{Invariance and Robustness}
\label{ssec:invariance}
\begin{table}[t]
\centering
\footnotesize
\caption{Invariance and noise robustness. Probes: lower $\downarrow$ = more invariant. Invar.: lower $\downarrow$ = more consistent. Noise: concentration higher $\uparrow$,\textbf{Noise}: normalised token-distribution entropy / ids used (Ent/Ids, lower $\downarrow$ is better) RMS SD lower $\downarrow$. \textbf{Bold}: best per column.}
\setlength{\tabcolsep}{3pt}

\label{tab:invariance}
\begin{tabular}{lrrrrrr}
\toprule
& \multicolumn{2}{c}{Probes (\%) $\downarrow$}
& \multicolumn{2}{c}{Invar. $\downarrow$}
& \multicolumn{2}{c}{Noise} \\
\cmidrule(lr){2-3} \cmidrule(lr){4-5} \cmidrule(lr){6-7}
Model & Spk & Emo & DTW & Edit & Ent/Ids$\downarrow$ & RMS SD$\downarrow$ \\
\midrule
\multicolumn{7}{l}{\textit{Baselines}} \\
HuBERT           & 93.1 & 55.4          & 0.1609          & 0.2275          & 0.499/139          & 0.9051 \\
WavLM            & 78.9 & 54.6          & 0.1218          & 0.2096          & 0.642/191          & 1.0010 \\
\midrule
PINT (ours) &  1.2 & 32.1          & 0.0092          & 0.0659          & \textbf{0.000/1} & \textbf{0.0049} \\
\midrule
\multicolumn{7}{l}{\textit{Ablations}} \\
dec-AR           & 20.1 & 44.2 & 0.0269 & 0.1654 & 0.470/153 & 0.7092 \\
synth-only       &  4.1 & 32.9 & 0.0089 & 0.2485 & 0.488/175 & 0.6055 \\
w/o noise        & \textbf{1.2} & \textbf{27.5} & 0.0370 & 0.1412 & 0.566/198 & 0.7527 \\
dec-CTC          & 43.0 &  43.1 & 0.0485 & 0.3176 & 0.329/65   & 0.4989   \\
synth-aug        &  2.3 & 32.3  & 0.0075 & 0.0869 & 0.000/1   & 0.0053   \\
\bottomrule
\end{tabular}
\end{table}

We evaluate invariance through probes, parallel-utterance consistency, and noise robustness.
\textbf{Speaker/emotion probes}: x-vector time-delay neural network (TDNN) classifiers~\cite{snyder2018xvector} on frozen representations; speaker identity on CSTR-VCTK held-out speakers, emotion+intensity on RAVDESS; lower accuracy indicates greater invariance.
\textbf{Parallel invariance} (CSTR-VCTK test set): DTW cosine ratio (mean intra-group DTW-aligned cosine distance normalized by inter-group distance) and average normalized edit distance between deduplicated token sequences of parallel utterances normalized by the longer sequences length.
\textbf{Noise robustness}: \emph{Entropy} for discrete: Entropy of the distribution of discrete id's when feeding 2 noise classes \cite{noise} not seen in training through the encoders, this is paired with the number of unique ids used atleast once; For continuous: \emph{RMS standard deviation ratio (RMS SD)} root-mean-square of per-dimension standard deviations for noise vs.\ speech embeddings (near zero = negligible noise variation).
Results are shown in Table~\ref{tab:invariance}.

PINT reduces speaker probe accuracy from 93\,\% (HuBERT) to 1.2\,\%, demonstrating near-complete speaker invariance. Emotion (7 emotions + neutral)  accuracy also drops substantially (55\,\%\,$\to$\,32--33\,\%), though some signal persists. Parallel-corpus DTW distances confirm that continuous representations collapse intra/inter group distance by ${\sim}20\times$ relative to HuBERT with a much lower sequence variablity also at the discrete level. Noise robustness is essentially perfect for PINT (entropy~$=0$, RMS SD~$<0.02$), whereas baselines exhibit substantial drift.
The ablations reveal that each component and the data contribute distinctly. Using sDTW without word level loss leads to collapse. The \emph{dec-AR} variant retains 20.1\,\% speaker accuracy which is ${\sim}16.8\times$ worse than full PINT. \emph{synth-only} achieves low speaker leakage but high edit distance (0.25 vs.\ 0.06), indicating that while synthetic variation removes speaker signal, atleast some real data is needed for discrete sequence consistency. \emph{w/o noise} achieves the lowest emotion leakage but poor noise concentration and high DTW distance, confirming that noise augmentation is important for robustness without sacrificing invariance.

\subsection{Compression Efficiency}
\label{ssec:compression}

When identical content consistently maps to identical (similar) tokens, sequences contain clear patterns that compression algorithms can exploit. We test this prediction with progressive compression schemes (Table~\ref{tab:compression_efficiency}), reporting tokens per second and bits per second ($\lceil\log_2 V\rceil \times$ tok/s). The \textbf{Tx} column applies the same formula to characters and character level BPE on transcripts, providing a text-grounded compressibility reference.
\begin{table}[t]
  \caption{Encoding compressibility: \textit{tok/s\,/\,b/s} per cell (lower~$\downarrow$ = more compact). \textbf{Hu}: HuBERT-L9; \textbf{WL}: WavLM-L12; \textbf{PI}: PINT-L12. \textbf{Tx}: character-level BPE on transcripts.}
  \label{tab:compression_efficiency}
  \centering
  \footnotesize
  \begin{tabular}{llcccc}
    \toprule
    & & \multicolumn{4}{c}{\textit{tok/s\,/\,b/s}} \\
    \cmidrule(lr){3-6}
    \textbf{Enc.} & $\boldsymbol{V}$ &
      \textbf{Hu} & \textbf{WL} & \textbf{PI} & \textbf{Tx} \\
    \midrule
    Raw       &    & 50/400   & 50/400   & 50/400            & 14.6/116 \\
    Dedup     &    & 24.6/197 & 24.8/198 & \textbf{12.6}/101 & ---      \\
    RLE       &    & 24.6/246 & 24.8/273 & \textbf{12.6/152} & ---      \\
    BPE-orig  & 4k & 15.0/180 & 15.3/183 & 10.0/120          & 4.1/50   \\
    BPE-orig  & 8k & 13.2/172 & 13.4/175 & 8.8/114           & 3.7/48   \\
    BPE-dedup & 4k & 9.3/112  & 9.4/113  & 4.9/59            & ---      \\
    BPE-dedup & 8k & 8.2/107  & 8.3/108  & \textbf{4.3/56}   & ---      \\
    \bottomrule
  \end{tabular}
\end{table}
PINT RLE achieves \textbf{152,bits/s}—a $2.6\times$ reduction from 400,bits/s—just 31,\% above raw text (116,bits/s), while retaining frame-level time alignment. HuBERT and WavLM reach only 246–273,bits/s (${\sim}1.6\times$ compression), as speaker and channel variability disrupt token runs with surface-level changes rather than true content shifts.

The deduplication rate (74.8\,\% for PINT vs.\ 50.7\,\% for HuBERT) is the most direct fingerprint of nuisance invariance: it teases out precisely the consistent structure that both RLE and BPE can exploit.  After deduplication and BPE at $V{=}8{,}192$, PINT reaches \textbf{56\,bits/s} approaching text BPE (48\,bits/s) while baselines stall at 107-108\,bits/s.  BPE merges token $n$-grams that recur consistently across the corpus; because PINT minimizes dependence on speaker and acoustic conditions, the same content more reliably maps to the same token sequence, and BPE discovers stable patterns just as it finds recurring character sequences in text. In HuBERT and WavLM, the same content uttered under different conditions produces different token sequences, fragmenting these patterns and blocking the merges.

\begin{figure}[t]
  \centering
  \includegraphics[width=\columnwidth]{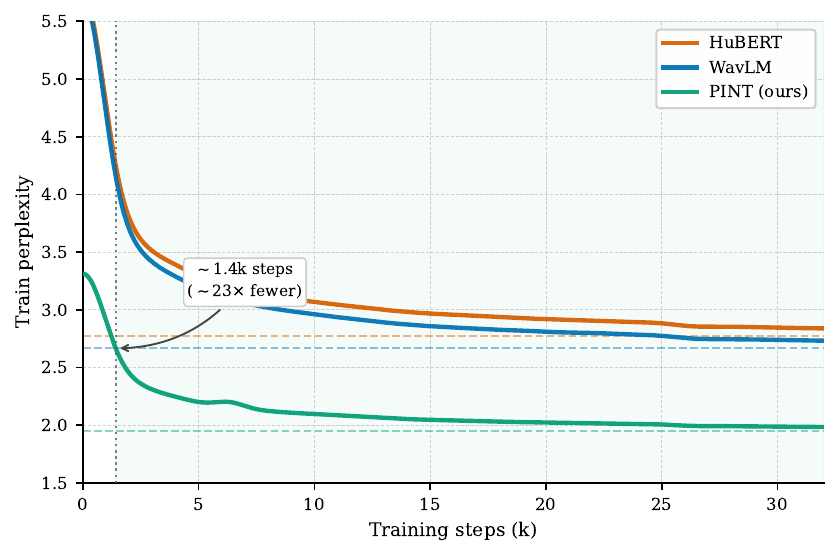}
  \caption{Training perplexity on LibriLight for an identical 85\,M-parameter transformer LM on HuBERT\,L9, WavLM\,L12, and PINT\,L12 ($k{=}200$) tokens. Dashed lines: final test perplexity. PINT converges to 27--30\,\% lower perplexity, consistent with its superior sequence compressibility (Table~\ref{tab:compression_efficiency}).}
  \label{fig:lm_perplexity}
\end{figure}

\noindent\textbf{Autoregressive LM perplexity.}
The ultimate test of token quality is whether an autoregressive model finds them easy to predict. We train identical 85\,M-parameter decoder-only Transformers (12 layers, hidden dimension 1024, 16 heads, context 3072 tokens, weight-tied embeddings) on raw token id  streams from each tokenizer using LibriLight. Figure~\ref{fig:lm_perplexity} shows the training curves. PINT achieves test perplexity \textbf{1.95}, versus 2.78 for HuBERT and 2.67 for WavLM, a \textbf{27--30\,\%} reduction under identical architecture and data. Strikingly, PINT already matches WavLM's  perplexity after only ${\sim}1{,}400$ steps, roughly \textbf{23$\times$ fewer iterations} and continues to improve far below the baseline ceiling. PINTs supperior token consistency directly reduces the burden that nuisance leakage forces on the LM.

\section{Conclusion}
PINT shows that enforcing invariance yields audio content compressibility close to text, providing a cleaner interface to NLP techniques. By treating parallel utterances as natural supervision and jointly satisfying invariance and discriminability, the resulting tokens are speech-grounded and speaker-agnostic, with a 27--30\,\% perplexity reduction confirming that nuisance leakage bottlenecks learning efficiency.
As invariant, frame-aligned representations, PINT tokens serve as drop-in targets for codec architectures and disentanglement-sensitive tasks. Future work includes multilingual extension via synthetic data, codec integration, and duration-factorized generation schemes that exploit PINT's RLE compressibility.
\newpage

\section{Generative AI Use Disclosure}
Some code used in the experiments was written with help from a
coding assistant (Claude by Anthropic). The experiments were run manually and results were manually verified. Generative AI
was also used in the formatting of tables and plots.
The paper was manually written. The authors assume full responsibility and
accountability for the content of this submission.

\bibliographystyle{IEEEtran}
\bibliography{mybib}

\end{document}